\documentclass[conference]{IEEEtran}
\IEEEoverridecommandlockouts
\usepackage{cite}
\usepackage{amsmath,amssymb,amsfonts}
\usepackage{algorithmic}
\usepackage{graphicx}
\usepackage{textcomp}
\usepackage{xcolor}
\usepackage{comment}
\usepackage{multirow}
\usepackage{pgf}
\usepackage{tabularx}
\usepackage{tikz}

\def\BibTeX{{\rm B\kern-.05em{\sc i\kern-.025em b}\kern-.08em
    T\kern-.1667em\lower.7ex\hbox{E}\kern-.125emX}}

\usepackage[square,numbers]{natbib}

\DeclareMathOperator*{\argmin}{arg\,min}

\newcommand\copyrighttext{%
  \footnotesize \textcopyright 2020 IEEE.  Personal use of this
  material is permitted.  Permission from IEEE must be obtained for
  all other uses, in any current or future media, including
  reprinting/republishing this material for advertising or promotional
  purposes, creating new collective works, for resale or
  redistribution to servers or lists, or reuse of any copyrighted
  component of this work in other works.}
\newcommand\copyrightnotice{%
\begin{tikzpicture}[remember picture,overlay]
\node[anchor=south,yshift=10pt] at (current page.south) {\fbox{\parbox{\dimexpr\textwidth-\fboxsep-\fboxrule\relax}{\copyrighttext}}};
\end{tikzpicture}%
}

\begin{document}

\title{Neural Entity Linking on Technical Service Tickets\\
}

\author{
\IEEEauthorblockN{1\textsuperscript{st} Nadja Kurz}
\IEEEauthorblockA{\textit{RheinMain University of} \\
\textit{Applied Sciences}\\
Wiesbaden, Germany \\
kurz.nadja@gmx.de}
\and
\IEEEauthorblockN{2\textsuperscript{nd} Felix Hamann}
\IEEEauthorblockA{\textit{RheinMain University of} \\
\textit{Applied Sciences}\\
Wiesbaden, Germany \\
felix.hamann@hs-rm.de}
\and
\IEEEauthorblockN{3\textsuperscript{rd} Adrian Ulges}
\IEEEauthorblockA{\textit{RheinMain University of} \\
\textit{Applied Sciences}\\
Wiesbaden, Germany \\
adrian.ulges@hs-rm.de}
}

\maketitle
\copyrightnotice

\begin{abstract}
  Entity linking, the task of mapping textual mentions to known
  entities, has recently been tackled using contextualized neural
  networks. We address the question whether these results --- reported
  for large, high-quality datasets such as Wikipedia --- transfer to
  practical business use cases, where labels are scarce, text is
  low-quality, and terminology is highly domain-specific.

  Using an entity linking model based on BERT, a popular transformer
  network in natural language processing, we show that a neural
  approach outperforms and complements hand-coded heuristics, with
  improvements of about 20\% top-1 accuracy. Also, the benefits of
  transfer learning on a large corpus are demonstrated, while
  fine-tuning proves difficult. Finally, we compare different
  BERT-based architectures and show that a simple sentence-wise
  encoding (Bi-Encoder) offers a fast yet efficient search in
  practice.
\end{abstract}

\begin{IEEEkeywords}
  Entity Linking, Attention Models, Natural Language Processing
\end{IEEEkeywords}

\section{Introduction}
Entity Linking refers to the challenge of matching entity mentions in
text (e.g. ``FAZ'') with the corresponding entity they refer to
(e.g. \textit{Frankfurter\_Allgemeine\_Zeitung}). The task is an
essential building block to various NLP tasks such as information
extraction~\cite{jurafsky17nlp} or question
answering~\cite{dubey18qa}. While some challenges (such as spelling
errors or abbreviations) can be resolved using heuristics, more
complicated cases (such as synonyms, hyponyms or coreferences) must be
resolved based on mentions' contexts. Here, recent neural models based
on representation learning on large-scale text collections (typically
using language modeling) have achieved impressive results and are
commonly considered state-of-the-art~\cite{vaswani2017attention,
  devlin18bert}.

These results have mostly been reported for web-based encyclopedic
text (namely, Wikipedia) and knowledge graphs (namely, Freebase). On
the other hand, information extraction is increasingly applied for
knowledge management in business contexts. Even in vast sectors with a
rather technical focus such as mechanical engineering, service and
other knowledge based activities are becoming more and more important
to revenue (``service is the new marketing''). Given the rapid
development of technology paired with a scarce and increasingly
fluctuating workforce, experience management becomes vital: domain
experts and service technicians must be enabled on the job even when
inexperienced or when facing outdated technology. Frequently,
documentation is scarce and reduces to hastily collected error
messages, service reports, chat protocols or CRM entries. Linking this
information to leverage it for a highly efficient support is a vital
challenge to knowledge centered businesses. It requires the semantic
indexing of unstructured text collections, from the classification of
terms and phrases over the automatic extraction of facts to
dialogue-based interfaces for digital assistants.

In this paper, we address the question whether the results achieved
with neural models on large web-based encyclopedic text and knowledge
graphs can be transferred to entity linking ``in the wild'' (more
precisely, in a business context). We present a study in cooperation
with our business partner Empolis Information Management, where
automated entity linking is of vital interest to facilitate a highly
efficient knowledge engineering on a diverse landscape of customer
data. Particular, our study focuses on entity linking in a mechanical
engineering application scenario: Entities refer to machine parts such
as ``Flansch''/``Luftanschluss'' (flange) or error symptoms such as
``Leck''/``Ölaustritt'' (leakage). About 1200 of such entities are to
be linked to mentions in over 1.6 million tickets collected by service
technicians in the field. This setting differs substantially from
Wikipedia-based scenarios:

\begin{itemize}
    \item {\it Domain Change}: While Wikipedia covers a breadth of topics, business contexts come with particular terminology, entities and relations. This raises the question how a domain-specific model compares to a generic one pre-trained on Wikipedia.
    \item {\it Density and Quality}: The German Wikipedia contains over 2 million entities, each with concise textual descriptions and manually annotated mentions (backlinks). Business domains are often less rich and come with little annotated data. Also, texts such as error diagnoses and repair descriptions --- particularly when entered via mobile devices --- tend to suffer from spelling errors and incorrect grammar.
    \item {\it Language}: Finally, business texts come in arbitrary languages (in our case, German). This raises the question whether language-specific models or generic multi-lingual models~\cite{devlin18bert} are preferable.
\end{itemize}

We explore these aspects using different entity linkers based on the
state-of-the-art transformer network BERT~\cite{devlin18bert}. The
approach is based on context matching, i.e. a match likeliness is
derived from {\it sentences} containing the entity mention on the one
hand and the reference entity on the other.  Particularly, we study
different BERT-based model architectures and pre-training strategies
on Wikipedia excerpts as well as the business-domain tickets mentioned
above. Our findings are:

\begin{itemize}
    \item The BERT model does offer benefits compared to a rule-based baseline using manually engineered heuristics. Particularly, combining both approaches in a simple hybrid model gives strong improvements on all domains.
    \item We found pre-training to be vital, and fine-tuning on the target domain's limited data to be of little use.
    \item We compare two BERT-based approaches~\cite{humeau2019poly}, either encoding \textit{single} sentences (Bi-Encoder) or sentence {\it pairs} (Cross-Encoder, which allows attention between sentences). While the latter comes with small performance improvements, the first approach is significantly more scalable and thus preferable in practice.
\end{itemize}

\section{Related Work}
Research on entity linking (or {\it named entity
  disambiguation})~\cite{jurafsky17nlp, ji2011knowledge} is usually
conducted on Wikipedia-based datasets, where hyperlinks between
wikipages serve as mentions and the link targets as
entities~\cite{mihalcea2007wikify, milne2008learning}. We distinguish
between \textit{graph-}, \textit{embedding-} and
\textit{attention-based} approaches:

\textbf{1. Graph-based:} These methods aim to encode the relationships
that entities form with each other as features to exploit possible
interdependencies between them. A method to encode pair-wise
interdependence is proposed by \citet{milne2008learning}. Here models
are trained to classify the proposed entity based on a set of
statistical features including \textit{relatedness} (which encodes the
semantic proximity of wikipages based on their shared
hyperlinks). \Citet{han2011collective} extend this approach by
defining a weighted graph where vertices are both entities+mentions,
and perform a Page-Rank like inference.

\textbf{2. Embedding-based:} Embedding based approaches map entities,
mentions and sometimes mention contexts to dense vector
representations. \Citet{sun2015modeling} propose neural tensor
networks~\cite{socher2013reasoning} as encoders, trained to minimize
the cosine distance between correct entity mention
pairs. \Citet{yamada2016joint} use the previously mentioned
relatedness feature to learn entity embeddings jointly with context
and mention embeddings to obtain a geometrical
alignment. \Citet{francis2016capturing} train a model which learns the
similarity between a mention and an entity by applying a logistic
regression to a feature matrix comprised of both statistical
information gathered on Wikipedia and a feature map of a CNN that
received word embeddings of the mention. \Citet{nguyen2016joint}
follow a similar approach but use RNNs to model both local and global
features of the documents containing the mentions and learn to rank
candidate entities by their likelihood to be the correct one. Finally,
\Citet{gupta2017entity} train a joint model to produce entity
embeddings from a combination of mentions, contexts and
descriptions. The most likely entity is determined by using the
mention-context-encoder of the joint model.

\textbf{3. Attention-based:} Most similar to our work are
embedding-based methods that rely on the {\it attention mechanism},
where embeddings are contextualized using an attention-based
architecture such as transformer
networks~\cite{devlin18bert}. \Citet{yamada2019pre} approach the
problem by fine-tuning a pre-trained transformer model
\cite{vaswani2017attention}. They optimize a masked entity prediction
task where the model is trained to predict an entity given an entity
and a sample sentence containing the entity. \Citet{kolitsas2018end}
use a bidirectional sentence encoder and combine it with the attention
mechanism proposed by \citet{ganea2017deep} to obtain a score for
mention-entity pairs while considering a given textual context both
locally (on sentence level) and globally (considering all other
mentions on document level). \Citet{logeswaran2019zero} and
\citet{wu2019zero} study the problem of entity linking in a zero-shot
scenario where the model is confronted with entities unseen in
training (we also tackle this setting). \Citet{logeswaran2019zero}
generate entity candidates by selecting mentions via BM25 scoring and
then re-rank those with different BERT-based models. Both encoders
score with the dot product of their output. \Citet{wu2019zero} employ
BERT-based models called {\it Bi-Encoder} and {\it Cross-Encoder} for
ranking. We evaluate similar architectures in our work. In contrast to
this, our approach neither needs special encodings nor markers for the
Bi-Encoder and can thus be used without any fine-tuning.

\section{Approach}

We assume a set of entities $ E = \{e_1, e_2, \dots, e_m \} $ to be
given (such as \textit{New\_York\_Times}).  Entity linking is targeted
at mapping a textual entity mention (such as ``Times'') to the correct
entity. Each mention comes with a context sentence (such as ``The {\it
  Times} reported the Dow Jones to drop by $1$\%'').

The entity linker is {\it trained} on a different set of entities $E'$
with $E' \cap E {=} \emptyset$ (i.e., we tackle an open-world setting
where entity linking is applied to novel entities unseen when
learning). 
For {\it all} entities $e \in E {\cup} E'$, we assume a set of
\textit{reference sentences} $S(e)$ to be given, each containing a
mention of the entity.

We evaluate three entity linking models: a heuristic one, a BERT-based
one, and a hybrid method combining the other two. Those strategies are
outlined in the following.

\subsection{Heuristic Linking}
\label{heuristic}

A simple way of matching a mention (such as ``FAZ'') to an entity
(such as \textit{Frankfurter\_Allgemeine\_Zeitung}) is to simplify
both strings 
and then perform a string comparison. Our first approach follows this
strategy and applies a set of symbolic transformations (or {\it
  heuristics}) $ f $ on the given mention $ m_i $ and entity name
$e_j$ to find matches of the form $ f(m_i) \approx f(e_j) $. Thereby,
we compare $f(m_i)$ and $f(e_j)$ with the Damerau Levenshtein distance
$ d^L(f(m_i), f(e_j))$~\cite{boytsov2011indexing}.  We adopted several
heuristics commonly used in name matching tasks, and picked the
following combination $f_1,...,f_7$ which worked best on our
Wikipedia-based datasets:

 \begin{enumerate}
 \item \textbf{Remove Punctuation:} removes non-alphanumeric
   characters such as !,\#,(,),[, or ].
 \item \textbf{Corporate Forms:} removes corporate suffixes,
   e.g. \textit{HolidayCheck Group AG} $ \to $
   \textit{HolidayCheck}.
 \item \textbf{Lowercasing}, for example \textit{IBM} $ \to $
   \textit{ibm}
 \item \textbf{Stemming}, for example \textit{working} $ \to $
   \textit{work}
 \item \textbf{Stopword} removal, e.g. \textit{Procter and Gamble}
   $ \to $ \textit{Procter Gamble}
 \item \textbf{Sorting} tokens alphanumerically,
   e.g. \textit{reeves keanu} $ \to $ \textit{keanu reeves}.
 \item \textbf{Abbreviations} All token n-grams are abbreviated to
   their initials, after decomposing words into their compounds
   \footnote{https://github.com/dtuggener/CharSplit},
   e.g. \textit{allgemeiner wirtschaftsdienst} $ \to $
   \textit{awd}.
 \end{enumerate}

 For inference, a yet unlinked mention $ m $ and entity $e$ are
 transformed by applying each heuristic to the outcome of its
 predecessor. We then define the distance between $e$ and $m$ as the
 minimum Levenshtein distance over the sequence of heuristics:

$$
\begin{aligned}
  d(m,e) := min_{t=0,...,7} \;\; d^L  \;\Big(& \,\, f_t \circ \dots \circ \, f_1 \, (m), \\
  & \,\, f_t \circ \dots \circ f_1 \, (e) \;\,\,\, \Big)
\end{aligned}
$$

Finally, the heuristic linker returns the entity $e$ with minimum
distance $d(m,e)$ to mention $m$. A scalable search over all entities
is achieved by indexing with SymSpell
\footnote{https://github.com/wolfgarbe/SymSpell}.

\subsection{Contextual Linking}
\label{bertlinking}
\begin{figure}[t]
  \includegraphics[width=\columnwidth]{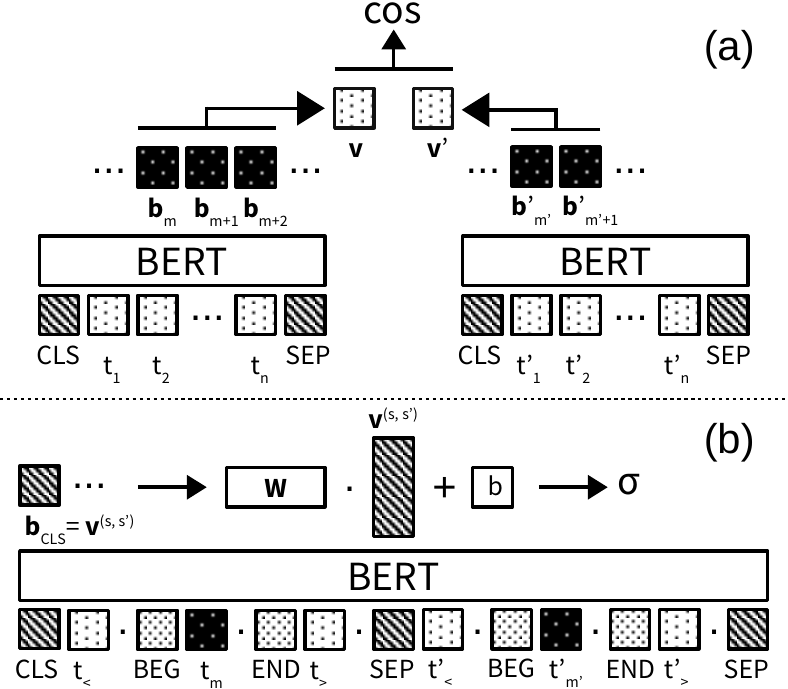}
  \caption{Model Overview: On top is the Bi-Encoder (a) model where a
    BERT is fine-tuned in a Siamese fashion to reduce the distance of
    the pooled mention embeddings. Below is the Cross-Encoder (b)
    version where two sentences are fed into BERT
    simultaneously. Here, the mentions are wrapped by special BEG and
    END tokens. The CLS embedding is regressed to a scalar to learn
    the likelihood of the two mentions expressing the same entity.}
  \label{fig:model_overview}
\end{figure}

While the heuristic linker is based on a string comparison of mention
and entity name, our second method includes mentions' context
sentences.  The approach matches mentions in query sentences with
mentions of known entities in reference sentences. All mentions are mapped to dense, contextualized vectors ({\it embeddings}). For inference, the entity with the closest reference sentence compared to the mention's context is returned. We investigate two approaches employing BERT as an encoder, referred to as {\it Bi-Encoder} and {\it Cross-Encoder} in reminiscence to~\citet{humeau2019poly}. Both approaches (illustrated in Figure \ref{fig:model_overview}) fine-tune BERT on the entity linking task by aligning mentions referring to the same entity.\\

\noindent \textbf{Bi-Encoder:} Let
$ s = ( t_1, \dots\;, t_m, \dots\;, t_{m+k},$ $ \dots\;, t_n) $ be a
sentence in which the (word-piece) token subsequence
$t_m, \dots\, t_{m+k}$ refers to the entity mention. BERT transforms
$s$ to an embedding sequence $\mathbf{b}_1, \dots\, \mathbf{b}_{n}$ of
$768$-dimensional vectors (excluding the obligatory CLS and SEP
tokens). We define the mention's representation as the average over
its tokens' BERT embeddings:
\begin{equation}
\label{eq:avg-embedding}
\mathbf{v}^{(s)} = \frac{1}{k+1} \cdot  (\mathbf{b}_m + \dots + \mathbf{b}_{m+k})
\end{equation}

In training, we sample pairs $(s,s')$ of sentences either describing
the same entity (positive samples) or not (negative samples). These
sentences are then transformed by the model and their cosine distance
\( \text{d}(s,s') := 1 {-} cos(\mathbf{v}^{(s)}, \mathbf{v}^{(s')}) \)
is computed.  The (BERT) model's weights are updated using SGD with a
contrastive max-margin loss:

\begin{equation}
L^{(\text{mm})}(s, s', y) := y \cdot \text{d}(s, s')^2 + (1{-}y) \: \text{max}(\gamma {-} \text{d}(s, s'), 0)^2
\label{eq:bi_encoder_loss}
\end{equation}

where the label $y \in \{0,1\}$ denotes whether a sentence pair is
positive (1) or negative (0), and $\gamma$ is the margin.

For inference, the sentence containing the mention to be linked is
matched against each entity $e$'s set of reference sentences
$S(e)$. The entity with the most similar BERT mention embedding is
returned:

\begin{equation}
    e^*(s) \; := \; 
    \argmin_{e \in E} \;\, \min_{s' \in S(e)} \: \text{d}(s, s'). 
\end{equation}

A scalable search over all entities is realized using approximate nearest neighbor search with annoy\footnote{https://github.com/spotify/annoy}.\\

\noindent \textbf{Cross-Encoder} In the second approach the sentence
pairs are not successively but simultaneously passed to the
model. This enables attention between the sentences, which has been
proven beneficial in other tasks~\cite{humeau2019poly}.  To do so,
special markers are introduced before/after the beginning/end of a
mention. Furthermore, the input sentences $s,s'$ are separated by a
dedicated separator token (see Figure \ref{fig:model_overview}).

Instead of averaging token embeddings as in Equation
\eqref{eq:avg-embedding}, we define the mention pair's representation
$\mathbf{v}^{(s,s')} := \mathbf{b}_{\text{CLS}} $ to be BERT's
embedding of the classifier token. From this, we infer the probability
that the two mentions in their respective context refer to the same
entity:

\begin{equation}
    P(s,s') := \sigma \big( \mathbf{W} \cdot  \mathbf{v}^{(s,s')} + b \big),
\end{equation}

where \( \mathbf{W} \in \mathbb{R}^{d \times 1} \) reduces the BERT
CLS representation to a scalar and $\sigma$ denotes the sigmoid
activation function.  As a training objective, we use a binary cross
entropy loss:

\begin{equation}
  \begin{split}
  L^{(\text{ce})}(s,s',y) :=  -\Big( & y \cdot \log(P(s,s')) \\
    & + (1 {-} y) \cdot \log(1 - P(s,s'))\Big)
  \end{split}
\label{eq:cross_encoder_loss}
\end{equation}

Note that --- since both sentences are passed simultaneously through
BERT --- this model does {\it not} allow to precompute embeddings in
an index structure but requires a sequential pairwise comparison of
the input sentence with all reference samples.

\subsection{Hybrid Linking}
It is reasonable to assume that heuristic matching offers a reliable
solution for simple cases such as abbreviations or spelling errors,
while the BERT-based linker can disambiguate more complicated cases
based on context. Therefore, we combine both using a simple strategy:
We first apply the heuristic linker. If this returns {\it none} or
{\it multiple} suitable entities with the same distance, the
BERT-based linker is applied instead.

\section{Experiments}
For our experiments we investigate the advantages of contextual
linking approaches over heuristic approaches as well as the
combination of both methods.  We further investigate the importance of
the language in which a neural model is trained on, and how well these
methods can be applied to {\it synonym detection}, a practical
knowledge engineering use case.

\subsection{Data}
We evaluate our entity linking methods using two structured,
high-quality Wikipedia-based datasets as well as the business domain
dataset, which contains a significant amount of colloquialism and
spelling errors.

\noindent \textbf{Wikipedia -- Mixed+Geräte:} We crawled two different
datasets (\textit{Geräte} and \textit{Mixed}) from the German
Wikipedia using the internal hyperlink structure that links words or
phrases between articles. For our approach, each article is considered
an entity and referred to by multiple mentions in different contexts.
The Mixed dataset contains diverse entities, while --- as a topical
approximation to the engineering target domain --- the {\it Geräte}
dataset focuses on entities of the category \textit{appliances}.

\noindent \textbf{Domain-specific -- Empolis:} Our third dataset,
\textit{Empolis}, represents our business use case, and contains
domain-specific entities and corresponding synonyms (such as ``Leck'',
``Pickup-Säge'' or ``Motorflansch'') commonly used in our business
partner's ticket corpus. For our contextual linking approach, we use
the tickets in which the synonyms appear as context information.

An overview of the size of all datasets is given in Table
\ref{tab:dataset}.  The training split is used to fine-tune the neural
models, the validation split for tuning hyperparameters and the
testing split for evaluation.  All three splits contain disjoint sets
of entities, i.e. we test on different entities than
validating/training on.  Furthermore, the sentences are divided into
reference and query sentences ($50{/}50$\% for Wikipedia, $30{/}70$\%
for Empolis).

\begin{table}[]
    \caption{Dataset statistics (number of entities and sentences)}
    \centering
    \begin{center}
        \begin{tabular}{l|rr|rr|rr}
        \multirow{2}{*}{ Dataset } & \multicolumn{2}{c}{ Train } & \multicolumn{2}{c}{ Test } & \multicolumn{2}{c}{ Validation } \\
        & \#Ent. & \#Sent. & \#Ent. & \#Sent. & \#Ent. & \#Sent. \\
        \hline
        Mixed & 8331 & 107082 & 1027 & 12853 & 1031 & 13560 \\
        Geräte & 5717 & 65101 & 698 & 7680 & 3231 & 35823 \\
        Empolis & 401 & 13587 & 200 & 6601 & 201 & 6281 \\
    \end{tabular}
    \end{center}
    \label{tab:dataset}
    \vspace{-2.3em}
\end{table}

\subsection{Hyperparameters}

We evaluate the contextual linking approach using different BERT
models: We start with a pre-trained, multilingual off-the-shelf BERT
model\footnote{https://github.com/google-research/bert} (orig).  Later
experiments include variations which have been fine-tuned on the
entity linking task using the losses in Equation
\ref{eq:bi_encoder_loss} and \ref{eq:cross_encoder_loss}. Finally, we
also test a domain-specific BERT created by fine-tuning a pre-trained
German BERT model\footnote{https://deepset.ai/german-bert} on the
Empolis data using the standard masked language modelling loss.

\subsection{Heuristic vs Contextual Linking}

We compare the top-1 accuracy of the heuristic linking approach with
the BERT-based Bi-Encoder in Table \ref{tab:heuristic_contextual}.
Additionally, we evaluate how well the combination of the heuristic
linking approach with the Bi-Encoder (\textit{Hybrid}) performs.  For
the Bi-Encoder approaches, results using the original multilingual
BERT model as well as a fine-tuned version are reported.

\begin{table}[h]
    \caption{Comparing different linking approaches on two Wikipedia excerpts (Geräte, Mixed) and the partner's business data (Empolis). A combination of heuristic and BERT-based approaches (Hybrid) offers the highest accuracy.}
    \begin{center}
    \begin{tabular}{l|ccc}
        \multirow{2}{*}{ Classifier } & \multicolumn{3}{c}{ Top1 Accuracy [\%] } \\
        & Geräte & Mixed & Empolis \\
        \hline
        Heuristic & 77.87 & 83.98 & 51.16 \\
        \hline
        BERT-Bi (orig) & 93.30 & 95.93 & 40.06 \\
        BERT-Bi (fine-tuned) & 93.32 & 97.25 & 35.11 \\
        \hline
        Hybrid (orig) & \bf 94.72 & 97.52 & \bf 71.40 \\
        Hybrid (fine-tuned) & 93.40 & \bf 97.84 & 69.76 \\
    \end{tabular}
    \end{center}
    \label{tab:heuristic_contextual}
\end{table}


The results show that the combination of heuristic linking with
contextual information ({\it Hybrid}) leads to the best
performance. This improvement is most significant on the Empolis
target domain where we observe improvements of the linking accuracy of
up to 20\% relative to the heuristic baseline. Here, we found the
models to best complement each other (heuristic linking covers simple
cases reliably, BERT-based linking more complicated synonyms). A few
entity linking examples using our hybrid approach are detailed in
Table \ref{tab:entity_linking_examples}.

\begin{table}[h]
\caption{ Examples of the entity linking results achieved using our hybrid approach. Query mentions are highlighted in bold}
\begin{center}
    \begin{tabularx}{0.48\textwidth}{X|r}
         Query & Linked Entity \\
         \hline
         Der Kunde hat erheblichen \textbf{Ölaustritt} direkt an der Frässpindel. & Leck \\
         \hline
         Die \textbf{Flanschscheibe} hat starke Riefen von dem defekten Freilauf. & Flansch \\
         \hline
         Nach dieser \textbf{Blütezeit} geriet Naturasphalt über Jahrunderte hinweg in Vergessenheit. & Goldenes\_Zeitalter \\
         \hline
         Koch entdeckte den \textbf{Choleraerreger} sowie die Überträger von Pest und Malaria. & Vibrio\_cholerae
    \end{tabularx}
    \end{center}
    \label{tab:entity_linking_examples}
    \vspace{-1.6em}
\end{table}

\subsection{Bi-Encoder vs Cross-Encoder}

Next, we address the question which BERT-based architecture is more
suitable. Table \ref{tab:be_vs_ce} compares Bi-Encoder and
Cross-Encoder. Both approaches are evaluated individually as well as
in combination with the heuristic linking approach
(\textit{Hybrid}). Note that --- due to the introduction of additional
markers --- the Cross-Encoder can only be evaluated using a fine-tuned
BERT model.

While the Bi-Encoder offers a highly scalable search via index
structures, the Cross-Encoder requires pairs of samples to be
processed by the BERT model. Correspondingly, the time needed to
determine an entity is proportional to the amount of reference samples
in a dataset. This increases the processing time rapidly. Therefore,
we conduct the experiment only on a reduced number of randomly sampled
queries of about 700, 1000 and 200 mentions for the \textit{Geräte},
\textit{Mixed} and \textit{Empolis} dataset respectively.

\begin{table}[h]
\caption{Comparing the different BERT models BERT-Bi vs. BERT-Cross (reduced number of test samples)}
\begin{center}    \begin{tabular}{l|ccc}
        \multirow{2}{*}{ Classifier } & \multicolumn{3}{c}{ Top1 Accuracy [\%] } \\
        & Geräte & Mixed & Empolis \\
        \hline
        BERT-Bi (orig) & 89.68 & 93.09 & \bf 51.53 \\
        BERT-Bi (fine-tuned) & 90.40 & 96.01 & 43.88 \\
        BERT-Cross (fine-tuned) & \bf 94.13 & \bf 96.88 & 45.41 \\
        \hline
        Hybrid (BERT-Bi, orig) & 93.41 & 97.08 & 80.61 \\
        Hybrid (BERT-Bi, fine-tuned) & 93.27 & 97.37 & 77.04 \\
        Hybrid (BERT-Cross, fine-tuned) & \bf 96.42 & \bf 98.05 & \bf 81.63 \\
    \end{tabular}
    \end{center}
    \label{tab:be_vs_ce}
\end{table}

Comparing the Bi- and Cross-Encoder, we can see that while the
Cross-Encoder does perform better on all datasets, the practical
performance increase for the Hybrid-Model is limited ($1{-}4\%$). In
contrast, the Bi-Encoder is considerably faster by a factor of $1500$
(\textit{Geräte}), $3500$ (\textit{Mixed}) and $100$
(\textit{Empolis}) --- Comparing a single query sentence against all
6601 reference sentences takes around 23 seconds with the
Cross-Encoder. These results highlight the effectiveness of the
Bi-Encoder in practical settings.

\subsection{Domain vs Multilingual BERT}

Next, we tackle the issue of switching from the large, generic
Wikipedia domain to the specific business use case. We compare the
performance of the pre-trained multilingual BERT model with a
domain-specific BERT model (German BPE tokens, pre-trained on the
domain using language modeling).  We further evaluate the performance
of both models after fine-tuning them on the entity linking task.
Similar to the previous experiment the evaluation has been performed
using a reduced number of samples.

\begin{table}[h]
\caption{Performance of the multilingual BERT (right) vs a
  domain-specific BERT Model (left) \newline (reduced number of test samples)}
    \centering
    \begin{tabular}{l|ccc|ccc}
        \multirow{3}{*}{ Model } & \multicolumn{3}{c}{ Domain BERT } & \multicolumn{3}{c}{ Multilingual BERT } \\
        & \multicolumn{3}{c}{ Top1 Accuracy [\%] } & \multicolumn{3}{c}{ Top1 Accuracy [\%] } \\
        & Geräte & Mixed & Empolis & Geräte & Mixed & Empolis \\
        \hline
        Bi (orig) & 83.81 & 84.13 & 44.90 & 89.68 & 93.09 & 51.53 \\
        Bi (fine) & 88.54 & 94.26 & 38.27 & 90.40 & 96.01 & 43.88 \\
        Cross & 93.41 & 96.01 & 45.92 & 94.13 & 96.88 & 45.41 \\
    \end{tabular}
    \label{tab:domain_vs_multilingual_bert}
\end{table}

Overall, Table \ref{tab:domain_vs_multilingual_bert} shows that the
domain-specific BERT model (left) performs worse than the generic one
(right). In contrast to the Wikipedia data --- where fine-tuning
proves beneficial --- on the Empolis data neither a language model
fine-tuning nor a supervised finetuning on the entity linking task
were found to give improvements. Obviously, our target domain behaves
very differently to Wikipedia datasets. We attribute this difference
to the difficulty of the task on less structured (and lower quantity)
data.

\subsection{Synonym Discovery on Empolis}

Finally, we evaluate the utility of the Bi-Encoder in a practical use
case: The detection of synonyms for domain concepts (such as
``Ölaustritt'' and ``Leck'') is key to understanding unstructured
domain specific text, e.g. to identify descriptions of a reoccurring
problems. In practice, synonyms are acquired by experts during the
knowledge engineering process, which requires domain expertise and is
extremely time-consuming. Therefore, it is of practical interest to
suggest synonyms automatically.

Given test entities on the Empolis dataset, we apply our model (using
the Hybrid BERT Bi-Encoder) to identify synonyms in the Empolis corpus
and return a ranked list of suggestions. Especially, we evaluate how
well our model can identify {\it new} synonyms which have been {\it
  missed} by human experts. In order to determine synonyms of a query
entity, we use a PoS-Tagger \footnote{https://spacy.io (model:
  de\_core\_news\_sm)} to detect nouns and link each noun to an entity
using its context sentence. If the resulting entity is identical to
the query entity, the noun is stored as potential synonym with its
respective distance.

\begin{figure}[h]
\centering
\resizebox{0.45\textwidth}{!}{
\input{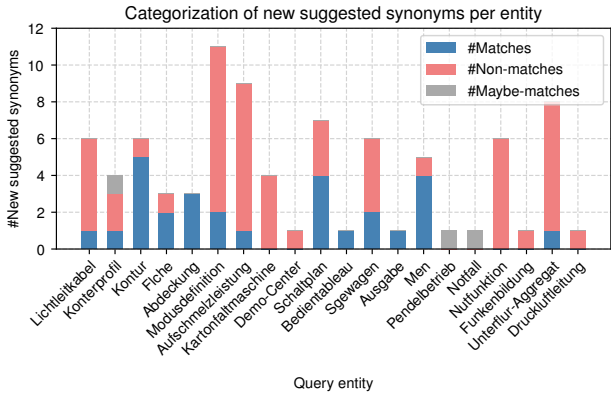}
}
\caption[New synonyms suggested by the hybrid classifier.]{New synonyms suggested by the Hybrid model on the Empolis data.}
\label{fig:empolis_unknown_synonyms}
\vspace{-1.3em}
\end{figure}

We collected synonym suggestions for $20$ query entities and labeled
their correctness manually into three categories (matches,
non-matches, maybe-matches).  Results are detailed in Figure
\ref{fig:empolis_unknown_synonyms}. We observe that our system is
definitely helpful in a practical knowledge engineering setting,
suggesting at least one correct synonym (blue) in $13$ of $20$ cases
and achieving an average precision of $35$\%. Additionally to the
known synonyms acquired by experts, the Bi-Encoder is able to identify
$1.4$ new correct synonyms per entity on average.

\section{Discussion}

In this paper, we have studied the application of state-of-the-art
deep language models for entity linking in a business context. We show
that an ensemble of symbolic transformations and a neural approach
using BERT achieves impressive results --- both on Wikipedia excerpts
and a very noisy real-world dataset from an industry partner. It can
be highlighted that our Bi-Encoder approach, which offers the
opportunity to cache a set of pre-computed reference samples, is
performing comparably to an expensive Cross-Encoder approach. This
enables a resource-efficient production use-case.  While a
straightforward fine-tuning to the target domain fails so far,
developing effective strategies of fine-tuning to limited noisy
domains will be our main focus in the
future.

\section*{Acknowledgment}

This work was funded by German Federal Ministry of Education and
Research (Program FHprofUnt, Project DeepCA (13FH011PX6)).

\bibliography{sds20}
\bibliographystyle{plainnat}  

\end{document}